\documentclass[12pt,a4paper]{article}
\usepackage[utf8]{inputenc}
\usepackage[english]{babel}
\usepackage{geometry}
\usepackage{booktabs}
\usepackage{longtable}
\usepackage{array}
\usepackage{xcolor}
\usepackage{listings}
\usepackage{hyperref}
\usepackage{graphicx}

\title{Token-Oriented Object Notation vs JSON: A Benchmark of Plain and Constrained Decoding Generation}
\author{Ivan Matveev \\
  \texttt{i.a.matveev@gmail.com}}
\date{February 2026}

\usepackage{url}

\begin{document}

\maketitle
\section{Summary}
Recently presented \textbf{Token-Oriented Object Notation (TOON)} aims to replace JSON as a serialization format designed for passing structured data to Large Language Models with significantly reduced token usage. While showing solid accuracy in LLM comprehension\footnote{\url{https://github.com/toon-format/toon}}, there is a lack of tests against JSON generation. Though never present in training data, TOON syntax is simple enough to suggest solid one-shot in-context learning could support accurate generation. The inevitable example prompt overhead can be an acceptable trade-off for shorter completions, especially in cases of multiple repair loops that also happen with JSON generation.
To test this, we conducted a simple benchmark, creating several test cases with regard to structural complexity, a validation pipeline, and comparing plain JSON generation vs structured output (SO) (constrained decoding) JSON generation vs TOON one-shot in-context learning generation. JSON SO was included to add a possible minimum token budget baseline into generations and also to set a starting point for the possibility of another experiment where the TOON vocabulary is provided to an xgrammar-like state machine to test TOON inference enforcement.
\vspace{0.5em}
\newline
\textbf{Key findings are the following}: TOON shows promising accuracy/token consumption ratio for in-domain generation tasks, though this advantage is often reduced by the "prompt tax" of instructional overhead in shorter contexts. Meanwhile, plain JSON generation shows the best one-shot and final accuracy, even compared with constrained decoding structured output, where the only significant advantage is the lowest token usage as a trade-off for slightly decreased accuracy overall and significant degradation for some models. Notably, for simple structures, this "lowest token usage" of constrained decoding outperformed even TOON. This result also hints that TOON enforcing via frameworks such as xgrammar may not yield the desired results.Furthermore, the results suggest a scaling hypothesis: TOON's true efficiency potential likely follows a non-linear curve, shining only beyond a specific point where the cumulative syntax savings of large datasets amortize the initial prompt overhead, though this may introduce new risks regarding indentation drift over long context windows.
\section{Objective}

Compare the effectiveness of three output formats for structured LLM generation across multiple models:
\begin{itemize}
  \item \textbf{JSON (J)} - plain JSON generation
  \item \textbf{JSON-SO (JSO)} - JSON with Structured Output (constrained decoding)
  \item \textbf{TOON (T)} - TOON format
\end{itemize}

\textbf{Key metrics:}

\begin{itemize}
  \item \textbf{One-shot accuracy} - percentage of correct outputs on first attempt
  \item \textbf{Final accuracy} -  percentage of correct outputs after all repair cycles
  \item \textbf{Token budget} - total tokens (prompt + completion) spent to achieve correct result
\end{itemize}

\section{Benchmark design}
\textbf{Reproducibility:} The complete source code, including dataset generators, prompt templates, and raw evaluation logs, is available in the public repository: \url{https://github.com/vetertann/TOON-generation-benchmark}.

\vspace{1em}
\textbf{Gold standard:} created from Pydantic models and serialized to \texttt{*.gold.json} (canonical JSON) and \texttt{*.gold.toon} (via \texttt{@toon-format/cli}\footnote{\url{https://github.com/toon-format/toon?tab=readme-ov-file##cli}}).

\vspace{1em}
\textbf{Test cases:}

\begin{enumerate}
  \item \textbf{users} - simple tabular structure
  \item \textbf{order} - nested structure with array
  \item \textbf{company} - department and employee hierarchy
  \item \textbf{invoice}  -items and totals
\end{enumerate}

\textbf{Test tracks:}

\begin{itemize}

  \item \textbf{JSON track (J)}: plain JSON generation with Pydantic validation
  \item \textbf{JSON-SO track (JSO)}: structured output (\texttt{response\_format="json\_object"}\footnote{\url{https://tokenfactory.nebius.com/}}) with constrained decoding. The model enforces valid JSON generation through constrained (guided) decoding, where it masks logits so that only tokens consistent with the defined grammar remain available at each step. The inference engine compiles these constraints (such as JSON Schema, regex, or grammar) into a state machine (e.g., xgrammar \cite{dong2024xgrammar}) and prunes illegal next tokens during generation, ensuring the model produces well-formed and syntactically correct outputs
  \item \textbf{TOON track (T)}: TOON output, CLI decoding, and validation using the same schemas 
\end{itemize}

Prompts for TOON contained universal and not per-case examples to make a fair comparison with JSON generation without custom-tailored schema example.
For comparison stability, canonicalization is applied. Each model was tested \textbf{10 times}, with results saved to CSV.

\vspace{1em}
\textbf{Sampling parameters:} 
All models were evaluated using 
temperature 0 to ensure deterministic (as much as it is possible) outputs. 
Each model was run 10 times per test case to verify output stability 
and measure consistency across API calls (not sampling variance). Other parameters were set 
to provider defaults.

\vspace{1em}
\textbf{Evaluation process}

\begin{enumerate}
  \item Prompt generation: LLM produces J, JSO, or T output. \textbf{21 models} provided via Nebius API \footnote{\url{https://tokenfactory.nebius.com/}}
  \item For TOON: CLI decoding to JSON. CLI errors trigger \textbf{repair cycle}
  \item Validation of all formats via Pydantic
  \item Data canonicalization
  \item Comparison with gold standard JSON
  \item If data doesn't match, a \textbf{repair cycle} starts (up to 3 attempts), where the previous output and error text are inserted into the prompt
\end{enumerate}

For each model run, one-shot and final accuracy are computed directly from
Boolean success results across the four evaluation cases (users, order,
company, invoice). One-shot accuracy is the fraction of cases solved on the
first attempt, while final accuracy is the fraction solved after up to three
repair cycles. Token usage is recorded at full case-level granularity
(prompt and completion tokens for each case and format). Per-format totals
(JSON, JSON-SO, TOON) are defined as the absolute sum of all prompt and
completion tokens consumed across the four cases in a run. We do not compute
a separate one-shot token budget: when a case succeeds on the first attempt,
its one-shot and final token costs coincide, while repair cycles add to the
final token total.

\section{Example prompts}

\subsection{JSON prompt (J \& JSO)}
{\small
\begin{lstlisting}
Create an order record:
- Order ID: 101
- Customer: Ada (ID: 9)
- Items:
  * Product A1: quantity 2, price $9.99 each
  * Product B2: quantity 1, price $14.50 each

Return as JSON with fields for id, customer (with id and name), 
and items array (with sku, qty, price).
\end{lstlisting}
}

\subsection{TOON prompt (T)}
{\small
\begin{lstlisting}
You are to produce output STRICTLY in TOON format.
        TOON RULES:
        - Use 2-space indentation
        - Scalars: fieldName: value
        - Objects: fieldName: then nested fields indented
        - Arrays of objects:
            arrayName[N]:
              - field1: value1
                field2: value2
        - Tabular arrays (for simple data):
            arrayName[N]{field1,field2}:
              val1,val2
              val3,val4
        - [N] MUST equal actual row/item count
        - Output ONLY a ```toon code block
        Reference example:
        ```toon
        id: 100
        type: Sample
        metadata:
          version: 1
          author: Alex
        sections[2]:
          - code: A
            title: Introduction
            items[2]{id,value}:
              1,First
              2,Second
          - code: B
            title: Details
            items[1]{id,value}:
              3,Third
        summary:
          total: 3
          status: complete
        ```
        TASK:
        Create an order record with fields: id, customer (with id and name), 
        and items array (with sku, qty, price).
        - Order ID: 101
        - Customer: Ada (ID: 9)
        - Items:
          * Product A1: quantity 2, price $9.99 each
          * Product B2: quantity 1, price $14.50 each

\end{lstlisting}
}
\section{Average results by model}

The per-model results are computed by grouping all evaluation runs by model and averaging the run-level aggregates. 
Each individual run already contains one-shot accuracy, final accuracy, and total token consumption, each of which is computed across all four test cases. 
Thus, the per-model table reports the mean of these run-level values for each model.

{
\footnotesize  
\setlength\tabcolsep{2.5pt}  
\begin{longtable}{@{}p{4.2cm} r r r r r r r r r@{}}
\toprule
\textbf{Model} & \multicolumn{3}{c}{\textbf{J}} & \multicolumn{3}{c}{\textbf{JSO}} & \multicolumn{3}{c}{\textbf{T}} \\
\cmidrule(lr){2-4} \cmidrule(lr){5-7} \cmidrule(lr){8-10}
 & \textbf{1-S} & \textbf{Fin} & \textbf{Tok} & \textbf{1-S} & \textbf{Fin} & \textbf{Tok} & \textbf{1-S} & \textbf{Fin} & \textbf{Tok} \\
\midrule
\endfirsthead

\toprule
\textbf{Model} & \multicolumn{3}{c}{\textbf{J}} & \multicolumn{3}{c}{\textbf{JSO}} & \multicolumn{3}{c}{\textbf{T}} \\
\cmidrule(lr){2-4} \cmidrule(lr){5-7} \cmidrule(lr){8-10}
 & \textbf{1-S} & \textbf{Fin} & \textbf{Tok} & \textbf{1-S} & \textbf{Fin} & \textbf{Tok} & \textbf{1-S} & \textbf{Fin} & \textbf{Tok} \\
\midrule
\endhead

\midrule
\multicolumn{10}{r}{\textit{Continued on next page}} \\
\endfoot

\bottomrule
\endlastfoot

NousResearch/Hermes-4-405B & 92.5\% & 92.5\% & 3252 & 35.0\% & \textbf{100\%} & 4759 & 50.0\% & 60.0\% & 4671 \\
NousResearch/Hermes-4-70B & 75.0\% & 75.0\% & 4414 & 37.5\% & 75.0\% & 5594 & 50.0\% & 50.0\% & 4738 \\
PrimeIntellect/INTELLECT-3 & 72.5\% & 75.0\% & 10682 & 72.5\% & 77.5\% & 10103 & 40.0\% & 65.0\% & 13315 \\
Qwen/Qwen2.5-Coder-7B-fast & 0.0\% & 0.0\% & 37705 & 75.0\% & 75.0\% & 4440 & 27.5\% & 27.5\% & 32715 \\
Qwen/Qwen3-235B-A22B-Inst & \textbf{100\%} & \textbf{100\%} & 2772 & \textbf{100\%} & \textbf{100\%} & 2772 & 50.0\% & \textbf{100\%} & 4715 \\
Qwen/Qwen3-235B-A22B-Thk & 82.5\% & 82.5\% & 11425 & 87.5\% & 97.5\% & 7899 & 50.0\% & 97.5\% & 17457 \\
Qwen/Qwen3-30B-A3B-Inst & 75.0\% & 75.0\% & 4436 & 75.0\% & 75.0\% & 4436 & 50.0\% & 70.0\% & 5505 \\
Qwen/Qwen3-32B & 75.0\% & 77.5\% & 10196 & 75.0\% & 75.0\% & 4120 & 47.5\% & 80.0\% & 9101 \\
Qwen/Qwen3-Coder-30B-A3B & 75.0\% & 75.0\% & 4206 & 75.0\% & 75.0\% & 4206 & 50.0\% & \textbf{100\%} & 4719 \\
Qwen/Qwen3-Coder-480B & 75.0\% & 75.0\% & 4462 & 75.0\% & 75.0\% & 4447 & 50.0\% & 75.0\% & 4515 \\
deepseek-ai/DeepSeek-R1 & 55.0\% & 70.0\% & 13811 & 65.0\% & 80.0\% & 4149 & 25.0\% & 50.0\% & 19047 \\
deepseek-ai/DeepSeek-V3-fast & 75.0\% & \textbf{100\%} & 3600 & 75.0\% & \textbf{100\%} & 3584 & 25.0\% & 80.0\% & 4734 \\
google/gemma-2-2b-it & 75.0\% & \textbf{100\%} & 4721 & 77.5\% & \textbf{100\%} & 4566 & 0.0\% & 0.0\% & 5955 \\
google/gemma-2-9b-it-fast & 75.0\% & 75.0\% & 6086 & 75.0\% & 75.0\% & 6056 & 50.0\% & 75.0\% & 5419 \\
meta-llama/Llama-3.3-70B & 75.0\% & 75.0\% & 4551 & 75.0\% & 75.0\% & 4447 & 50.0\% & 50.0\% & 5148 \\
meta-llama/Llama-3.1-8B & 72.5\% & 72.5\% & 7235 & 75.0\% & 75.0\% & 6941 & 22.5\% & 25.0\% & 4915 \\
moonshotai/Kimi-K2-Instruct & 50.0\% & 75.0\% & 4284 & 50.0\% & 75.0\% & 4283 & 50.0\% & \textbf{100\%} & 3937 \\
nvidia/Llama-3\_1-Nemotron & 75.0\% & 75.0\% & 4426 & 50.0\% & 50.0\% & 5714 & 50.0\% & 82.5\% & 4368 \\
openai/gpt-oss-120b & \textbf{97.5\%} & \textbf{100\%} & 3685 & \textbf{100\%} & \textbf{100\%} & 3545 & 50.0\% & 87.5\% & 8223 \\
openai/gpt-oss-20b & 50.0\% & 72.5\% & 14943 & 50.0\% & 67.5\% & 15601 & 50.0\% & 90.0\% & 9678 \\
zai-org/GLM-4.5 & 75.0\% & 87.5\% & 9677 & 75.0\% & 92.5\% & 9135 & 27.5\% & 52.5\% & 8110 \\
\end{longtable}
}

\section{Average results by case}

For each case (users, order, company, invoice), all case-specific accuracy and token columns are averaged over the entire evaluation dataset, i.e., across all models and all runs. 
Prompt and completion token counts are summed to reconstruct the total token budget for each format (J, JSO, T). 
This table therefore reflects the intrinsic difficulty and token cost of each case type, rather than model performance, providing a case-level profile of accuracy and efficiency across the full benchmark.

\begin{center}
\begin{tabular}{@{}l r r r r r r r r r@{}}
\toprule
\textbf{Case} & \multicolumn{3}{c}{\textbf{J}} & \multicolumn{3}{c}{\textbf{JSO}} & \multicolumn{3}{c}{\textbf{T}} \\
\cmidrule(lr){2-4} \cmidrule(lr){5-7} \cmidrule(lr){8-10}
 & \textbf{1-S} & \textbf{Fin} & \textbf{Tok} & \textbf{1-S} & \textbf{Fin} & \textbf{Tok} & \textbf{1-S} & \textbf{Fin} & \textbf{Tok} \\
\midrule
\textbf{users} & 94.8\% & 94.8\% & 1078 & 92.9\% & \textbf{100\%} & 556 & \textbf{90.5\%} & 90.5\% & 840 \\
\textbf{order} & 81.9\% & 81.9\% & 1746 & 78.6\% & 83.3\% & 1255 & 74.3\% & 78.6\% & 1585 \\
\textbf{company} & 18.6\% & 43.8\% & 3575 & \textbf{21.9\%} & \textbf{48.1\%} & 2592 & 0.0\% & 48.6\% & 2567 \\
\textbf{invoice} & 90.0\% & 90.0\% & 1723 & 87.6\% & \textbf{95.2\%} & 1349 & 0.0\% & 52.4\% & 3626 \\
\bottomrule
\end{tabular}
\end{center}

\section{Analysis and conclusions}

\subsection{The constrained decoding trade-off}
The data presents an interesting behavior and dichotomy in the performance of Structured Output (JSO track) compared to plain JSON (J track).

\textbf{1. Accuracy boost for weaker models:}
For smaller/less capable models, constrained decoding acts as a safety net. The most striking example \texttt{Qwen/Qwen2.5-Coder-7B-fast}. This model failed (J) completely (0\% accuracy) while consuming a massive 37,705 tokens. When constraints were applied (JSO), it achieved 75\% accuracy with a stable 4,440 tokens. One can assume that grammar enforcing act more as a guardrail here.

\textbf{2. Limitation for strong models:}
On the other hand, for larger reasoning models, strict grammar constraints often degrade initial performance. \texttt{NousResearch/Hermes-4-405B} dropped from 92.5\% one-shot accuracy in plain JSON to just 35.0\% in JSO. While the model quickly (based on the amount of overall token consumption) recovered to 100\% after repair cycles, the initial constraints appeared to significantly interfere with the model natural probability distribution, forcing it into suboptimal token paths. This aligns with the observations in \cite{schall2025hidden}.

\subsection{TOON domain alignment and application:}
Current results show that TOON's performance while significant is strictly dictated by the topology of the data. We can categorize the test cases into two distinct groups:

\begin{itemize}
    \item \textbf{TOON-aligned (Users, Order, Invoice):} Scenarios involving tabular data, "transactional" documents (nested objects plus uniform array) - shallow object nesting. These are within the format's intended design space.\footnote{\url{https://toonformat.dev/guide/getting-started.html}}
    \item \textbf{Non-aligned (Company case):} Scenarios involving deep nesting, recursive structures, or non-uniform arrays.
\end{itemize}

\begin{center}
{
\footnotesize
\setlength\tabcolsep{4pt}
\begin{tabular}{@{}l r r r r@{}}
\toprule
\textbf{Scenario Group} & \textbf{Metric} & \textbf{JSON (J)} & \textbf{JSO} & \textbf{TOON (T)} \\
\midrule
\textbf{TOON-Aligned} & \textit{1-Shot Acc} & \textbf{88.9\%} & 86.4\% & 54.9\% \\
(users/order/invoice) & \textit{Final Acc} & \textbf{88.9\%} & \textbf{92.8\%} & 73.8\% \\
\midrule
\textbf{Non-Aligned} & \textit{1-Shot Acc} & 18.6\% & \textbf{21.9\%} & 0.0\% \\
(company) & \textit{Final Acc} & 43.8\% & \textbf{48.1\%} & 48.6\% \\
\bottomrule
\end{tabular}
}
\end{center}

\subsection{Per-model token efficiency across aligned and non-aligned cases}

To validate the structural effects identified in the previous section,
we compute token efficiency separately for the two scenario groups:
(1) aligned cases (users, order) and (2) non-aligned cases (company and added invoice as less aligned).

For each model and track (J, JSO, T), we average:
\begin{itemize}
    \item the final accuracy across the cases in the group, and
    \item the per-case token cost (prompt + completion).
\end{itemize}

Efficiency is then defined as final accuracy per 1000 tokens.  
This method treats each case equally and avoids domination by
token-heavy cases, in contrast to the per-model aggregate table earlier.

\begin{figure}[htbp]
    \centering
    \includegraphics[width=\textwidth]{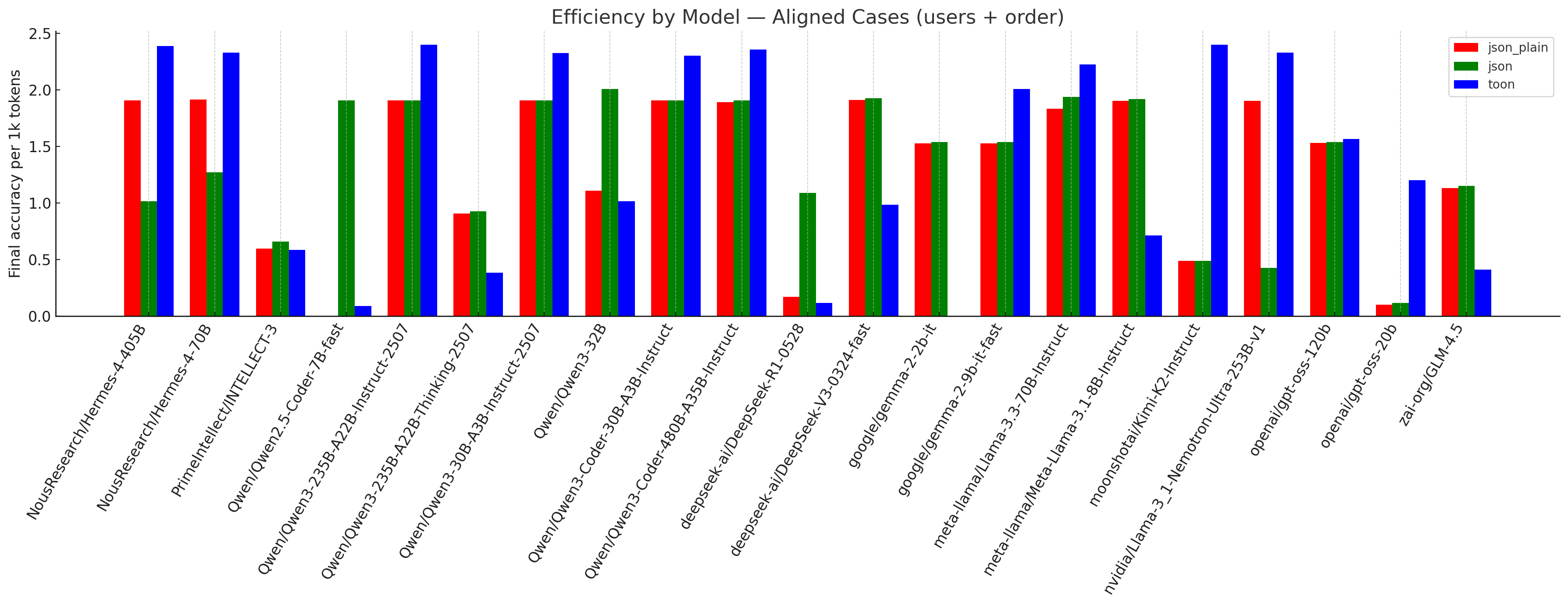}
    \caption{Efficiency by model for aligned cases (users + order).}
    \label{fig:aligned_efficiency}
    
    \vspace{1em}
    
    \includegraphics[width=\textwidth]{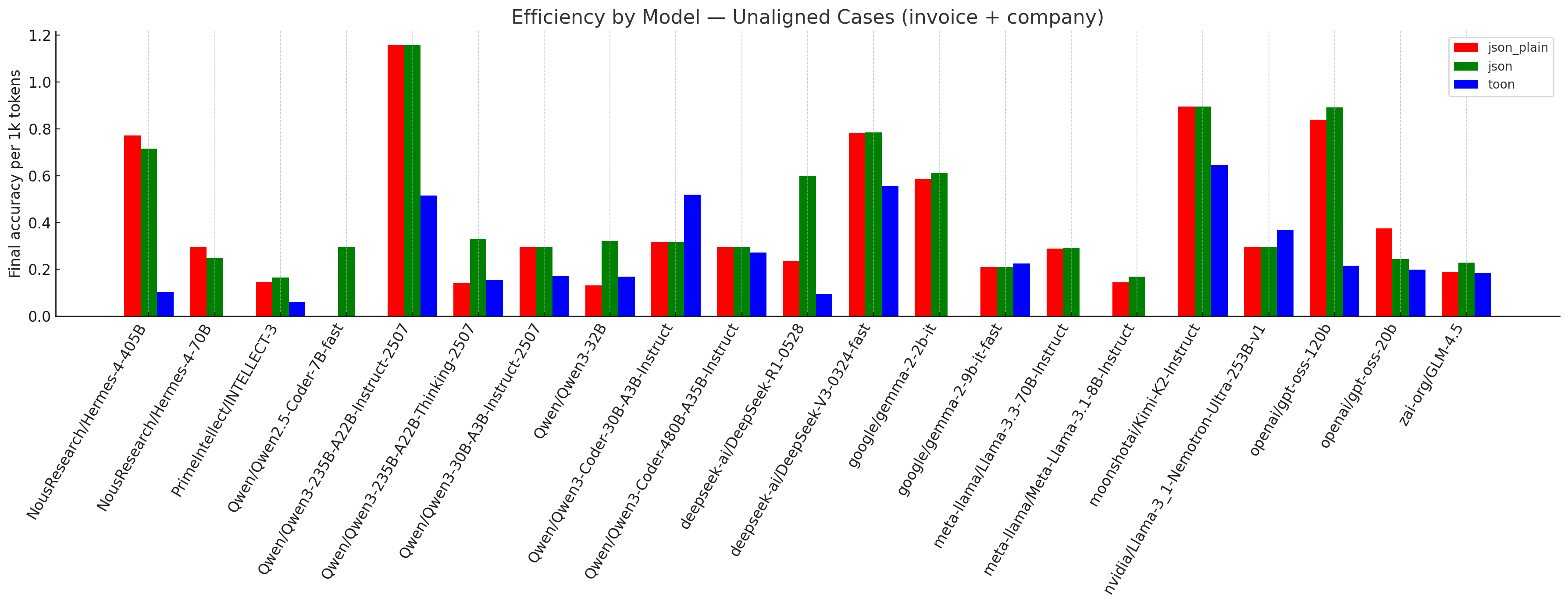}
    \caption{Efficiency by model for non-aligned cases (invoice + company).}
    \label{fig:unaligned_efficiency}
\end{figure}

These results quantify the intuition from the domain alignment analysis.
Models generally convert tokens into correct TOON generations more
efficiently in aligned scenarios, while performance collapses in
non-aligned cases, where JSON and JSON-SO retain their efficiency
advantage.

\textbf{Key Observations:}
\begin{enumerate}
    \item \textbf{The "sweet spot" (Tabular \& Nested+uniform):} TOON proved effective not just for flat arrays (\texttt{users}), but also for standard nested business entities. In the \texttt{order} case, which includes a parent object and a nested items array, TOON returned a solid \textbf{78.6\%} one-shot accuracy (comparable to JSO's 83.3\%). 
    
    \item \textbf{Efficiency in flat data:} In the \texttt{users} case, TOON achieved \textbf{90.5\%} one-shot accuracy while using \textbf{22\% fewer tokens} than plain JSON.
    
    \item \textbf{Non-aligned cases accuracy drop:} As structure moves from Aligned to Non-Aligned, TOON performance collapses. In the \texttt{company} case, the accuracy of \textbf{1-shot is 0\%}. Although repair cycles eventually brought the final accuracy to 48.6\% (comparable to JSON's 43.8\%, the cost in retries makes it inefficient for deep recursive hierarchies as it is. 
    But at the same time, Company case TOON final accuracy is the highest (48.6\% vs 43.8\% and 48.1\%), with the least amount of total token consumption (2567 vs 2592 and 3575). The reason is probably so massive syntactic overhead of such JSON structures in comparison with TOON that it pays off immediately. It therefore seems reasonable to assume that per‑case TOON examples that match schema better would probably reduce structural mistakes even further. Of course, data structure size scaling may result in non-linear error scaling, so this is up to additional research.

\end{enumerate}
\subsection{Token economy: the "prompt tax" and repair costs}

The results suggest a critical distinction between \textit{syntax efficiency} and \textit{inference efficiency}.

\textbf{1. The prompt tax:}
Standard JSON generation relies on the model's knowledge from train datasets, thus requiring minimal prompting. TOON as it is novel format, requires a heavy instructional prompt.
\begin{itemize}
    \item For many models (e.g., \texttt{Qwen/Qwen3-235B}), this overhead caused TOON to consume significantly more tokens (4715) than plain JSON (2772), despite the output format being lightweight.
    \item TOON only achieves a positive token trade-off when the generated output is large enough that it amortizes the fixed cost of the system prompt.
\end{itemize}

\textbf{2. Repair loop tax:}
Errors in generation lead to additional token costs. In the \textbf{invoice} case, the TOON track failed initially (0\% accuracy). Because the system prompt is large, feeding the error history back into the context for a repair loop caused token usage to double compared to JSON (3626 vs 1723). TOON generation is efficient when it works, but expensive when it fails.

\textbf{3. The JSO efficiency:}
For simple tabular data (\textbf{users} case), JSON-SO (Structured Output) proved to be the absolute token efficiency winner (556 tokens vs  TOON 840 tokens). Constrained decoding still might be a winner for simpler tasks.
\section{Recommendations}

\begin{enumerate}
    \item \textbf{Large-scale tests of TOON for "aligned" data streams:}
    TOON generation performance \textbf{may} result in significant performance boost in tabular and nested+uniform data structures, such as:
    \begin{itemize}
        \item SQL query result dumps (uniform tabular rows).
        \item Standard Transactional Documents (Orders, Invoices).
        \item Log streams or audit trails.
    \end{itemize}
    In these "aligned" cases, TOON offers a tangible reduction in token costs without sacrificing reliability (especially if one-shot example is customized for the task).

    \item \textbf{Avoid TOON for deep hierarchies:}
    For data representing complex state trees, DOM-like structures, or  any deeply nested configuration , TOON is NOT in any way production-ready. Use conventional serialization for these cases/ JSO if token limits are an issue.

    \item \textbf{Possible production strategy for LLMs:}
    Generation in TOON can be a valid choice for high-volume and cost-sensitive in-domain solutions, e.g. some ETL pipelines, batch processing, or scenarios involving heavy tabular/transactional data. If scales well, it can be production choice to minimize latency and token spend, provided the schema remains within the "aligned" domain.
    \item \textbf{Validate scaling and drift:}
    It is crucial to benchmark significantly larger dataset size where TOON's syntax savings effectively offset the prompt overhead. At the same time, TOON's reliance on whitespace may make it more susceptible to errors in long-context generations. 
\end{enumerate}

\bibliography{references}

\end{document}